\documentclass[runningheads]{llncs}
\usepackage{graphicx}
\usepackage{hyperref}
\usepackage{pdfpages}
\usepackage{tabularx}
\usepackage{multirow}
\usepackage{booktabs}
\usepackage{amsfonts}
\usepackage{amsmath}
\usepackage{tikz}
\usepackage{hyperref}
\usepackage{cite}
\usepackage{amsmath,amssymb,amsfonts}
\usepackage{graphicx}
\usepackage{textcomp}
\usepackage{subfig}
\usepackage{xcolor}
\usepackage{xspace}
\usepackage{bm}
\usepackage{adjustbox}
\usepackage{tabularx}
\usepackage{color}
\usepackage{algorithm}
\usepackage{algpseudocode}
\usepackage{listings}
\usepackage{nccmath}
\usepackage{subfig}
\usepackage{comment}
\usepackage[font=small, labelfont=bf]{caption}
\usepackage{mdwtab}
\usepackage{xspace}
\usetikzlibrary{fit}
\usetikzlibrary{positioning}
\usetikzlibrary{shapes.geometric, arrows}

\makeatletter
\DeclareRobustCommand\onedot{\futurelet\@let@token\@onedot}
\def\@onedot{\ifx\@let@token.\else.\null\fi\xspace}

\def\eg{\emph{e.g}\onedot} 
\def\ie{\emph{i.e}\onedot}

\makeatother

\begin{document}
\sloppy
\title{Spam-T5: Benchmarking Large Language Models for Few-Shot Email Spam Detection}
\titlerunning{Spam-T5: Benchmarking Large Language Models for Spam Detection}
%
\author{Maxime Labonne\thanks{Corresponding author} \and Sean Moran}

\authorrunning{Maxime Labonne \and Sean Moran}

\institute{JPMorgan Chase\\ \email{\{maxime.labonne,sean.j.moran\}@jpmchase.com}}
\maketitle              
\begin{abstract}
This paper investigates the effectiveness of large language models (LLMs) in email spam detection by comparing prominent models from three distinct families: BERT-like, Sentence Transformers, and Seq2Seq. Additionally, we examine well-established machine learning techniques for spam detection, such as Naïve Bayes and LightGBM, as baseline methods. We assess the performance of these models across four public datasets, utilizing different numbers of training samples (full training set and few-shot settings). Our findings reveal that, in the majority of cases, LLMs surpass the performance of the popular baseline techniques, particularly in few-shot scenarios. This adaptability renders LLMs uniquely suited to spam detection tasks, where labeled samples are limited in number and models require frequent updates. Additionally, we introduce \emph{Spam-T5}, a Flan-T5 model that has been specifically adapted and fine-tuned for the purpose of detecting email spam. Our results demonstrate that Spam-T5 surpasses baseline models and other LLMs in the majority of scenarios, particularly when there are a limited number of training samples available. Our code is publicly available at \url{https://github.com/jpmorganchase/llm-email-spam-detection}.
\end{abstract}
%
%
%


\section{Introduction}

Email communication continues to be an essential part of our daily lives, facilitating efficient asynchronous communication globally for personal and business users alike. Given this prominence, email is also a prime target for fraudulent and malicious activities such as spam and phishing attacks. Spam email can cause a multitude of problems ranging from user inconvenience arising from unsolicited communications, to overload of computational resources on servers and security compromise arising from fraudulent links and malware attachments in the emails that are designed to attack personal and business security. It is estimated that in 2022,
almost 49\% of emails sent over the internal were spam\footnote{\url{https://securelist.com/spam-phishing-scam-report-2022/108692/}}, highlighting the continued prevalence of the problem and the need to explore ever more sophisticated machine learning techniques to reduce the volume. Techniques for automatically detecting and filtering out spam emails are critical for enabling usability of personal and business email services and continue to attract significant research interest~\cite{Dada19}.

The detection of spam emails presents several challenges, which we categorize as \emph{data imbalance}, \emph{data distribution shift} and \emph{adversarial drift}. Despite the prevalence of email spam on the internet, one of the main obstacles in training spam detection models is the rarity of labeled datasets of fraudulent emails, which makes it difficult to obtain a representative sample for training effective machine learning models. This problem is further exacerbated for private, company-internal applications of email spam filtering, where there can be even less labeled data available due to the nature of the task (\eg~filtering out fraud and phishing attacks for specific user-groups such as high net worth individuals). This leads to the problem of \emph{imbalanced learning}, where the model may simply not have enough fraudulent samples to learn from, resulting in poor detection performance.

In addition, email communication is constantly evolving through time, providing an additional challenge. The changing nature of spam emails due to evolution and techniques used by spammers can lead to data distribution shifts in the dataset~\cite{Martino22,Tsymbal2004,Scholz2005AnEC,Syed99}. The problem of \emph{data shift} arises when the distribution of the training and test data is not the same, which can occur in real-world scenarios. For example, at a certain point in time, the word ``Casino'' could be indicative of spam, but the relative importance of this word might change (drift) through time. This violates the fundamental assumption of supervised learning and can cause classification models to fail to generalize in the deployment environment over time. Continual refresh of such models with recent representative data from the domain is critical. 

Furthermore, the environment is highly adversarial and characterized by a constantly developing arms race between spammers and email service owners, mainly driven by the lucrative gains that can be made through successful fraud and phishing attempts~\cite{Martino22}. Attackers have effectively created a cottage industry that constantly devises new and clever ways to bypass spam filters, resulting in an \emph{adversarial drift}. They focus their efforts on outmatching the textual filters by perturbing the data extracted from the email body and legible headers. Common strategies involve obfuscation techniques, disguising the content of the email, or manipulating the header information. 

To address these significant challenges, we claim that a promising approach is to use \emph{few-shot learning}~\cite{Wang20} to train classifiers that can detect fraudulent emails with limited samples. By using this approach, we can reduce the need for large labeled datasets and build classifiers that better generalize to unseen and constantly evolving data.

In this paper, we evaluate the performance of Large Language Models (LLMs) for sequence classification in a few-shot learning setting, compared to traditional machine learning techniques. Few-shot learning is ideal for the spam detection task in which the prevalence of the anomalous class (spam) is much less than the normal class (ham). Our main contribution is the development of a benchmark for traditional machine learning algorithms and LLMs on the four most popular datasets in spam detection. We evaluate the performance of these models in both a traditional supervised learning setting and a few-shot learning setting. Furthermore, we introduce a novel model, Spam-T5, which is a fine-tuned version of Flan-T5 specifically designed for email spam detection. Our findings show that Spam-T5 outperforms all other models on this benchmark. To facilitate further research in this area, we make our code and data publicly available at \url{https://github.com/jpmorganchase/emailspamdetection}.


\section{Related Work}
\subsection{Spam Detection using Machine Learning}

The email spam detection task has been well explored in the literature~\cite{Dada19} and is commonly used in undergraduate machine learning courses as an introductory use-case for study and learning. Despite the familiarity of the email spam detection task, it continues to provide real challenges to practitioners due to data distribution drift, the adversarial nature of the environment in which the task is embedded, and class imbalance. Significant research effort has been expended on this task, and sustained research is necessary to keep ahead of ever-evolving data and the changing landscape of spamming techniques.

The email spam detection task is framed as the development of an effective automated algorithm for differentiating spam versus ham (non-spam) emails. Spam detection methods have been classified in terms of rules-based, collaborative, and content-based approaches~\cite{Mendez06}. Rules-based approaches include checking incoming email addresses against white lists of allowed email addresses, black lists of commonly used spam email sources, and hand-crafted rules (\eg that look for empty \emph{To:} fields or a certain combination of keywords). Collaborative approaches compute, for example, a hash function on the content of example spam which is shared with the community and the hash function compared to new emails to detect spam~\cite{wittel04,Attenberg09,Shi11}. These simple methods are unable to generalize well and are brittle~\cite{Cranor98}, which is why content-based methods involving machine learning have been explored~\cite{Androutsopoulos2006LearningTF}. 

For content-based methods, early methods explored conventional machine learning methods and variations thereof, including Na\"ive Bayes classifiers~\cite{Sahami1998}, KNN~\cite{Delany05}, Support Vector Machines (SVMs)~\cite{cortes1995support,Torabi15,vapnik95,Drucker99}, and Boosting trees (\eg XGBoost)~\cite{Carreras01}. The task is frequently formulated as a binary classification problem where a classifier is trained on a representative dataset of spam and ham (non-spam) emails, and learns to classify the data points into these two classes. Performance is measured on generalization to new email data received, for example, on standard benchmark datasets~\cite{sakkis_memory-based_2003,sms_spam_collection,metsis2006naive} or in a live production operating environment for industrial use-cases. Studies have also been conducted on the effectiveness of various hand-crafted feature representations for input into the machine learning classifiers~\cite{Mendez06}. Other work has sought to tackle the concept drift problem, which is a defining aspect of the spam detection task~\cite{Scholz2005AnEC,Syed99}, and also to improve conventional classification techniques for the task~\cite{Alspector2001SVMbasedFO,hovold2005naive}. Generally speaking, conventional machine learning methods are faster to train and evaluate compared to deep neural networks, but have less modeling capacity to capture the complex patterns inherent in the data.

\subsection{Spam Detection using Large Language Models (LLMs)}

The field of machine learning has been revolutionized with the emergence of Large Language Models (LLMs), a powerful suite of deep learning methods that have exceptional abilities in understanding and learning from the structural, relational, and semantic patterns inherent in natural language datasets~\cite{liu2019roberta}. LLMs are extreme multi-taskers, able to replace most bespoke models for natural language tasks. For example they are very capable at text generation, sentiment detection, question answering, and text summarization. The release of ChatGPT and more recently GPT4~\cite{openai2023gpt4} to the public made waves around the world that continue to reverberate given the naturalness and human-like response from the model~\footnote{https://openai.com/blog/chatgpt, https://openai.com/research/gpt-4}. Despite the popularity and importance of email spam detection, there is little prior work that explores the benefits of LLMs for the task. We address the gap in this paper.

Given the robust popularity of the field, the literature on LLMs is many and varied, with new advances occurring rapidly and on daily basis. We focus on the suite of benchmark models that are commonly used as baselines in academic papers and in operation systems in industry, including RoBERTa~\cite{liu2019roberta},  SetFit~\cite{tunstall2022efficient}, and Flan-T5~\cite{weifinetuned}. These benchmark models have the advantage of publicly released code and weights, enabling comparison and evaluation on new tasks. Underlying almost all of the recent innovations in the field is the seminal Transformer architecture introduced by Google Research in 2017~\cite{Vaswani17}. The Transformer architecture dispenses with recurrent and convolutional layers, advocating the primary use of attention mechanisms for learning, which prove to be massively more scalable to big data. This technology was incorporated in the several generations (GPT-n) of Generative Pre-Trained Transformers (GPT) models~\cite{radford2018improving}, to spectacular effect across many natural language understanding tasks~\cite{openai2023gpt4}. Aside from the older GPT-2 model~\cite{Radford18}, the recent suite of GPT models are closed-source with weights hidden behind an API, making comparison impossible.

Among the open-source and widely available models are RoBERTa, SetFit and Flan-T5. BERT (Bidirectional Encoder Representations from Transformers)~\cite{devlin2019bert} addresses the issue of existing models only learning from previous tokens in text, creating more powerful (bi-directional) Transformer representations that gain from knowledge from an extended context. BERT has subsequently developed into a performant off-the-shelf model for many tasks. A subsequent, and popular evolution of BERT is embodied in the RoBERTa model~\cite{liu2019roberta} that improves BERT through a refined training methodology involving larger mini-batches and learning rates. SetFit (Sentence Transformer Fine Tuning)~\cite{tunstall2022efficient} is a recently proposed learning framework for efficiently (using orders of magnitude fewer parameters) fine-tuning sentence transformer models~\cite{reimers2019sentencebert}. Finally, and in the same spirit as SetFit, Flan-T5~\cite{chung2022scaling} is an improved model fine-tuning approach that leverages datasets of a larger number tasks framed as instructions and as applied to the T5 LLM~\cite{Raffel20}. The Flan fine-tuning methodology leads to significant gains on standard natural language understanding tasks. For the first time in the literature, we explore RoBERTa, SetFit, Flan-T5 and compare each to conventional models in this paper for the email spam detection task.

In terms of spam detection, some authors have applied early deep models to the task, including LSTM~\cite{HochSchm97} and BERT architectures~\cite{Debnath22}, although the literature currently is very sparse. We contribute to the field by exploring more recent LLM architectures in the few-shot learning paradigm. 


\section{Methods}

\subsection{Problem Formulation} Let $\mathcal{D} = \{(\mathbf{x}_i,y_i)\}^n$ be a dataset of emails, where each email $\mathbf{x}_i$ is represented as a feature vector in $\mathbb{R}^d$ and labeled as either spam ($y_i=1$) or not spam ($y_i=0$), also called ham. The spam detection classification problem can be formulated as learning a function $f: \mathbb{R}^d \rightarrow \{0,1\}$ that maps an email feature vector to a binary label, such that $f(\mathbf{x}_i)$ approximates $y_i$ for all email examples $i$ in the dataset.

This problem can be approached using supervised learning techniques, where a training set $\mathcal{D}_{train}$ is used to learn the function $f$. Specifically, the goal is to find the optimal parameters $\theta$ for a classifier $f_\theta$ that minimizes a suitable loss function $L(\theta)$, such as the cross-entropy loss:

\begin{equation}
L(\theta) = -\frac{1}{n}\sum_{i=1}^{n}(y_i\log(f_\theta(\mathbf{x}_i))+(1-y_i)\log(1-f_\theta(\mathbf{x}_i)))
\end{equation}

where $n$ is the number of emails in the training set. The optimal parameters $\theta^*$ can be found by minimizing the loss function:

\begin{equation}
\theta^* = \operatorname*{argmin}_\theta L(\theta)
\end{equation}

Once the optimal parameters are found, the learned classifier $f_{\theta^*}$ can be used to predict the spam label for new emails not seen during training.

However, in some scenarios, we may not have access to a large labeled dataset like $\mathcal{D}_{train}$. This is where few-shot learning comes into play. Few-shot learning is a variant of supervised learning that aims to learn from a small amount of labeled data~\cite{Wang20} In the few-shot learning setting, we are given a small support set $\mathcal{S}$ of labeled examples, where $\mathcal{S}=\{(\mathbf{x}_i,y_i)\}^k$, with $k\ll n$, and we need to learn a function $f$ that can generalize to new examples not seen during training.

This setting is particularly relevant in the context of spam detection, where labeled examples are scarce and require frequent updates to account for data shift and adversarial drift.

\subsection{Data Pipeline for Text Data}

Traditional text classification using machine learning involves preprocessing text data and extracting useful features. The resulting numerical features can be used to train machine learning models~\cite{survey_textclassification2,survey_textclassification}.

\begin{figure}[ht]
    \centering
    \definecolor{beige}{rgb}{0.9,0.96,1}

\tikzstyle{label} = [align=center]
\tikzstyle{node} = [align=center, fill=beige, rectangle, minimum width=2cm, minimum height=1cm, text centered, draw=black]
\tikzstyle{preprocess} = [rectangle, rounded corners, minimum width=7cm, minimum height=2.5cm, text centered, draw=black]
\tikzstyle{arrow} = [->,>=stealth]

\begin{tikzpicture}[node distance=2cm]
\node (text) [label] {Raw\\text};

\node (node1) [node, xshift=2.5cm] {Tokenize\\words};
\node (node2) [node, xshift=5cm] {Remove\\stopwords};
\node (node3) [node, xshift=7.5cm] {Stem or\\lemmatize};

\node (preprocessing) [preprocess, fit={(node1) (node2) (node3)}] {};
\node (label_preprocessing) [label, yshift=1.0cm, xshift=5cm] {Preprocessing};

\node (output) [label, xshift=10cm] {Preprocessed\\text};

\draw [arrow] (text) -- (preprocessing);
\draw [arrow] (node1) -- (node2);
\draw [arrow] (node2) -- (node3);
\draw [arrow] (preprocessing) -- (output);
\end{tikzpicture}
    \caption{Flowchart of the traditional preprocessing steps used for text classification. } \label{fig:flowchart}
\end{figure}
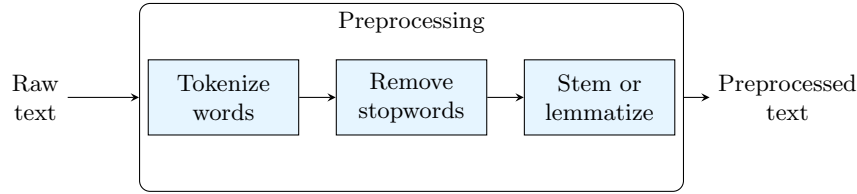

\subsubsection{Preprocessing.} The goal of preprocessing is to transform the raw text into a cleaner and more structured representation to apply a feature extraction algorithm (see Figure~\ref{fig:flowchart}).

The first step is \emph{tokenization}, which involves splitting the input text into individual words, phrases, or other units of meaning. The second step is to \emph{remove stopwords}, which are common words such as ``the", ``and", or ``in" that do not carry much semantic information. They can be safely discarded without affecting the meaning of the text. The third step is to apply \emph{stemming} or \emph{lemmatization}, which are techniques for reducing words to their base or root form. Stemming involves removing suffixes and prefixes from words, while lemmatization involves mapping words to their canonical or dictionary form.

Our preprocessing pipeline includes word tokenization and stemming, using the Porter stemming algorithm\footnote{\url{https://www.nltk.org/_modules/nltk/stem/porter.html}}~\cite{porter1980algorithm}.

\subsubsection{Feature Extraction.} The next step is to extract relevant features from the preprocessed text. Several algorithms can be considered for this task, such as bag-of-words~\cite{harris_1954} and Word2Vec~\cite{mikolov2013distributed}. In this study, we will focus on the popular Term Frequency-Inverse Document Frequency (tf-idf) encoding.

The tf-idf encoding is a common approach to representing text documents as numerical vectors, which can be fed to machine learning models. The main idea behind this encoding is to give a higher weight to words that are frequent in a \emph{particular} document (term frequency) but \emph{rare} in other documents (inverse document frequency). This helps to capture the unique characteristics of a document and distinguish it from other documents in the corpus.

Formally, the tf-idf encoding of a term $t$ in a document $d$ can be defined as:

\begin{equation}
\text{tf-idf}(t,d) = \text{tf}(t,d) \times \text{idf}(t)
\end{equation}

where $\text{tf}(t,d)$ is the term frequency of term $t$ in document $d$, and $\text{idf}(t)$ is the inverse document frequency of term $t$ calculated as:

\begin{equation}
\text{idf}(t) = \log\frac{N}{n_t}
\end{equation}

where $N$ is the total number of documents in the corpus, and $n_t$ is the number of documents that contain term $t$. The logarithmic function dampens the effect of rare terms with very low values of $n_t$.

\section{Experimental Setup}

\subsection{Datasets}

We leveraged four widely recognized spam datasets: the Ling-Spam dataset, SMS Spam Collection, SpamAssassin Public Corpus, and Enron Email dataset. These datasets were chosen for their popularity in the field of spam detection~\cite{agboola_spam_2022,sahmoud2022spam,tida2022universal} and the diversity of communication channels they represent, including SMS, mailing lists, and other sources.

\subsubsection{Ling-Spam Dataset.} The Ling-Spam dataset (2003) is a collection of messages used in experiments related to spam detection~\cite{sakkis_memory-based_2003}. The dataset is a mixture of spam messages and legitimate messages sent via the Linguist list, a moderated mailing list about linguistics. The corpus contains 2893 messages, with 2412 being legitimate messages obtained by randomly downloading digests from the list's archives and removing server-added text. The remaining 481 messages are spam messages received by one of the authors, translating to a spam rate of approximately 16\%. The Linguist messages cover various topics, including job postings and software availability announcements.

\subsubsection{SMS Spam Collection.} The SMS Spam Collection\footnote{\url{https://archive.ics.uci.edu/ml/datasets/sms+spam+collection}} (2011) is a dataset of 5,574 SMS messages in English that have been tagged as either ham or spam~\cite{sms_spam_collection}. The dataset was collected from various free or free-for-research sources on the internet. These sources include a UK forum where cell phone users report SMS spam messages, a dataset of legitimate messages collected for research at the National University of Singapore, a list of ham messages from a PhD thesis, and the SMS Spam Corpus v.0.1 Big. This is an imbalanced dataset with a spam rate of approximately 13\%.

\subsubsection{SpamAssassin Public Corpus.} The SpamAssassin dataset\footnote{\url{https://spamassassin.apache.org/old/publiccorpus/}} is a publicly available collection of email messages suitable for testing spam filtering systems. The dataset includes 6,047 messages, with a spam ratio of approximately 31\%. The messages were either posted to public forums, sent to the creator, or originated as newsletters from public news websites. The corpus is divided into five parts: spam, easy\_ham, hard\_ham, easy\_ham\_2, and spam\_2, with varying difficulty levels in differentiating spam from non-spam messages. In this study, we used the most recent versions of the five parts, from 2003 and 2005.

\subsubsection{Enron Email Dataset.} The Enron Email dataset\footnote{\url{https://www.cs.cmu.edu/~enron/}}, also known as the Enron Corpus, was collected in 2002 during the investigation into the bankruptcy of the Enron Corporation~\cite{metsis2006naive}. This dataset was generated by 158 employees and contains over 600,000 emails. We use a smaller version of this dataset, which contains 33,716 emails, labeled as spam or ham. The Enron Email dataset is balanced and contains 17,171 spam emails and 16,545 ham emails (spam rate of approximately 49\%).

\begin{figure}[ht]
\includegraphics[width=\textwidth]{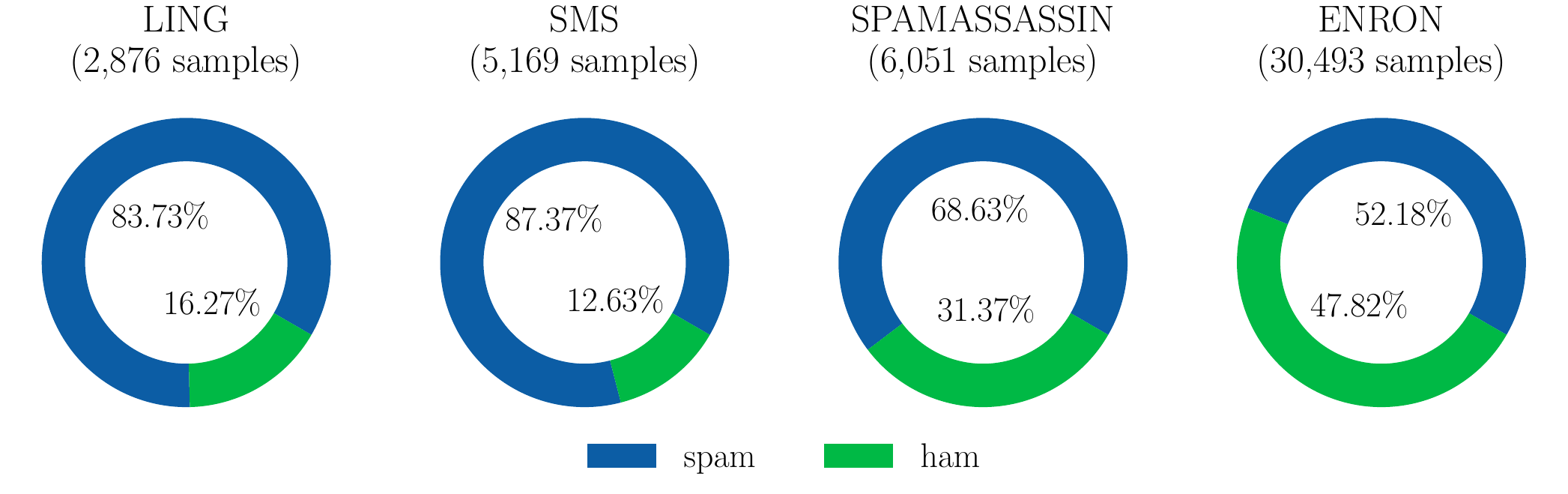}
\caption{Distribution of spam and ham messages across the datasets. Notably, the Ling, SMS, and SpamAssassin datasets exhibit an imbalanced learning scenario, where the prevalence of spam messages outweighs that of ham messages. Conversely, the Enron dataset is characterized by a balanced distribution of spam and ham messages.} \label{fig:piecharts}
\end{figure}

All these datasets were preprocessed by removing duplicates, NaN values, and empty messages. Figure~\ref{fig:piecharts} shows the proportions of spam and ham emails in every preprocessed dataset.

\subsection{Large Language Models}

We evaluate three large language models in this experiment (RoBERTa, SetFit, and Flan-T5) from three different families of architectures (BERT-like, Sentence Transformers, Seq2Seq).

\subsubsection{RoBERTa.} RoBERTa (2019) is a pretrained model for natural language processing~\cite{liu2019roberta} that builds on the success of BERT~\cite{devlin2019bert}. RoBERTa improves on BERT by modifying key hyperparameters, such as removing the next-sentence pretraining objective and training with larger mini-batches and learning rates. Additionally, RoBERTa explores training on a larger amount of data and for a longer period than BERT.

We used the HuggingFace implementation of the RoBERTa model (\url{roberta-base}) with the byte-level Byte-Pair Encoding (BPE) tokenizer~\cite{Radford18}.

\subsubsection{SetFit.} SetFit (2022) is an efficient and prompt-free framework for few-shot fine-tuning of Sentence Transformers (ST)~\cite{tunstall2022efficient,reimers2019sentencebert}. By fine-tuning a pretrained ST on a small number of text pairs in a contrastive Siamese manner, SetFit generates rich text embeddings that are used to train a classification head. This approach requires no prompts or verbalizers, and achieves high accuracy with significantly fewer parameters than existing techniques.

We used the SetFit implementation from the \texttt{setfit} library\footnote{\url{https://github.com/huggingface/setfit}}, combined with the sentence-transformers implementation of MPNet (\url{sentence-transformers/all-mpnet-base-v2})~\cite{song2020mpnet}, using the WordPiece tokenizer~\cite{wu2016googles}. Our implementation generates 20 training pairs. The distance between the resulting embeddings is measured using the cosine similarity.

\subsubsection{Flan-T5.} Flan-T5 (2022) is a family of models based on T5 (2019)~\cite{raffel2020exploring}, an encoder-decoder transformer architecture trained on multiple language tasks. The Flan-T5 models have undergone instruction-finetuning on over 1,800 language tasks, leading to a significant enhancement in their reasoning skills and promptability. However, it is worth noting that the Flan-T5 models were not trained to perform spam detection tasks.

We used the HuggingFace implementation of the Flan-T5 model (\url{google/Flan-t5-base}) with the SentencePiece tokenizer~\cite{kudo2018sentencepiece}. Our experimentation included the small version of the Flan-T5 model (80M parameters), but it demonstrated limited generalization capabilities, which is why it was excluded from this study.

The Flan-T5 model is a Seq2Seq model that is capable of generating textual outputs, as opposed to binary labels or probabilities. To leverage the capabilities of this model for spam detection, we fine-tuned it as a new task, introducing a dedicated prefix of ``\texttt{classify as ham or spam:}" to every sample. As a result, the model was trained to correctly output either ``\texttt{ham}" or ``\texttt{spam}" based on the input text. To obtain numerical values for classification metrics, a post-processing step was utilized to map the textual labels to $0$ and $1$. We call \emph{Spam-T5} this modified model fine-tuned on the spam detection task.

\subsubsection{Fine-tuning details.} We found that the batch size, learning rate, and number of epochs are all important hyperparameters that can affect how well the model generalizes to new data and how quickly it converges during training. Table~\ref{tab:parameters_llm} details the specific values used to fine-tune these models.

\begin{table}[ht]
\centering
\caption{Table captions should be placed above the tables.}\label{tab:parameters_llm}
\begin{tabular}{@{}l
>{\centering\arraybackslash}p{3.0cm}
>{\centering\arraybackslash}p{3.0cm}
>{\centering\arraybackslash}p{0.8cm}
>{\centering\arraybackslash}p{1.4cm}
@{}}
\toprule
\textbf{Model} & \textbf{Train batch size} & \textbf{Eval batch size} & \textbf{LR} & \textbf{Epochs} \\ \midrule
RoBERTa & 16 & 8  & 5e-5 & 10 \\
SetFit  & 16 & 16 & 2e-5 & 3  \\
Spam-T5 & 8  & 8  & 5e-5 & 5  \\ \bottomrule
\end{tabular}
\end{table}

\subsection{Baseline techniques}

We selected 6 baseline models that perform well in spam detection: Naïve Bayes (NB), Logistic Regression (LR), K-Nearest Neighbors (KNN), SVM, XGBoost, and LightGBM.

NB is a probabilistic model that assumes independence among features, making it fast and efficient for large datasets. LR is a linear model that uses a sigmoid function to predict binary outcomes. KNN is a non-parametric model that classifies data based on the proximity of its neighbors. In our implementation, we selected one neighbor ($K=1$). SVM is a linear model that finds the optimal hyperplane to separate data into different classes. We implemented it with a sigmoid kernel function and a gamma of $1.0$. XGBoost is a high-performing implementation of gradient boosting that utilizes a level-wise strategy to build decision trees in parallel and optimize the quality of splits in the training set. We set its learning rate to $0.01$ with 150 estimators. LightGBM is another gradient boosting framework that shares many of XGBoost's advantages, but differs in its leaf-wise construction of decision trees and use of a highly optimized histogram-based algorithm. Likewise, we set its learning rate to $0.01$ with 20 leaves.

\subsubsection{Fine-tuning details.} We found that the optimal number of tf-idf features was model-dependent. In order to tune this number, a stratified $5$-fold cross-validation technique was employed. This involved training the model with different numbers of tf-idf features and selecting the number that resulted in the highest performance. Table~\ref{tab:parameters_baseline} shows the optimal numbers of features for each model. 

\begin{table}[ht]
\centering
\caption{Number of features generated by the tf-idf algorithm for each model. These numbers were found using grid search on the validation set.}\label{tab:parameters_baseline}
\begin{tabular}{@{}l
>{\centering\arraybackslash}p{1.3cm}
>{\centering\arraybackslash}p{1.3cm}
>{\centering\arraybackslash}p{1.3cm}
>{\centering\arraybackslash}p{1.3cm}
>{\centering\arraybackslash}p{2.0cm}
>{\centering\arraybackslash}p{2.0cm}
@{}}
\toprule
\textbf{Model}       & \textbf{NB}   & \textbf{LR}  & \textbf{KNN} & \textbf{SVM}  & \textbf{XGBoost} & \textbf{LightGBM} \\ \midrule
\# features & 1000 & 500 & 150 & 3000 & 2000    & 3000     \\ \bottomrule
\end{tabular}
\end{table}


\section{Results}

\subsection{Full Training Sets}

We assess the performance of machine learning and large language models when they are provided with complete access to the training set. The complete training set in this context refers to 80\% of the entire dataset. To evaluate the performance of each model, we employ three metrics: F1 score (F1), precision (P), recall (R). The outcomes of the evaluation are presented in Table~\ref{tab:results_full}.

\begin{table}[t]
\centering
\caption{Test F1 score, precision, and recall performance of the six baselines and three LLMs for each dataset.}\label{tab:results_full}
\begin{tabular}{p{1.8cm}
>{\centering\arraybackslash}p{0.73cm}
>{\centering\arraybackslash}p{0.73cm}
>{\centering\arraybackslash}p{0.73cm}
>{\centering\arraybackslash}p{0.73cm}
>{\centering\arraybackslash}p{0.73cm}
>{\centering\arraybackslash}p{0.73cm}
>{\centering\arraybackslash}p{0.73cm}
>{\centering\arraybackslash}p{0.73cm}
>{\centering\arraybackslash}p{0.73cm}
>{\centering\arraybackslash}p{0.73cm}
>{\centering\arraybackslash}p{0.73cm}
>{\centering\arraybackslash}p{0.73cm}
}
\toprule
\textbf{} & \multicolumn{3}{c}{\textbf{Ling}} & \multicolumn{3}{c}{\textbf{SMS}} & \multicolumn{3}{c}{\textbf{SA}} & \multicolumn{3}{c}{\textbf{Enron}} \\ \cmidrule(l){2-4} \cmidrule(l){5-7} \cmidrule(l){8-10} \cmidrule(l){11-13}
\textbf{Model} &
  \multicolumn{1}{c}{\textbf{F1}} &
  \multicolumn{1}{c}{\textbf{P}} &
  \multicolumn{1}{c}{\textbf{R}} &
  \multicolumn{1}{c}{\textbf{F1}} &
  \multicolumn{1}{c}{\textbf{P}} &
  \multicolumn{1}{c}{\textbf{R}} &
  \multicolumn{1}{c}{\textbf{F1}} &
  \multicolumn{1}{c}{\textbf{P}} &
  \multicolumn{1}{c}{\textbf{R}} &
  \multicolumn{1}{c}{\textbf{F1}} &
  \multicolumn{1}{c}{\textbf{P}} &
  \multicolumn{1}{c}{\textbf{R}} \\ \midrule
NB           & \textbf{1.00} &       \textbf{1.00} &    \textbf{1.00} & 0.89 &       0.82 &    \textbf{0.98} & 0.87 &       0.83 &    0.91 & 0.96 &       0.96 &    0.96 \\
LR           & 0.98 &       0.96 &    \textbf{1.00} & 0.87 &       0.78 &    \textbf{0.98} & 0.92 &       0.89 &    0.96 & 0.97 &       0.98 &    0.96 \\
KNN          & 0.93 &       0.96 &    0.90 & 0.81 &       0.74 &    0.89 & 0.92 &       0.88 &    0.95 & 0.91 &       0.94 &    0.89 \\
SVM          & \textbf{1.00} &       \textbf{1.00} &    \textbf{1.00} & 0.90 &       0.83 &    \textbf{0.98} & 0.94 &       0.92 &    0.97 & 0.98 &       \textbf{0.99} &    0.97 \\
XGBoost      & 0.92 &       0.94 &    0.90 & 0.78 &       0.65 &    \textbf{0.98} & 0.94 &       0.92 &    0.96 & 0.91 &       0.98 &    0.85 \\
LightGBM     & 0.95 &       0.96 &    0.94 & 0.87 &       0.82 &    0.93 & \textbf{0.98} &       \textbf{0.98} &    \textbf{0.98} & 0.98 &       \textbf{0.99} &    0.96 \\ \midrule
RoBERTa      & 0.97 &       0.98 &    0.96 & 0.95 &       \textbf{0.97} &    0.92 & 0.97 &       0.98 &    0.95 & \textbf{0.99} &       \textbf{0.99} &    0.99 \\
SetFit & 0.99 &       0.98 &    \textbf{1.00} & \textbf{0.96} &       \textbf{0.97} &    0.95 & 0.95 &       0.96 &    0.94 &  \textbf{---} &  \textbf{---} & \textbf{---} \\
Spam-T5 & 0.99 &       0.98 &    \textbf{1.00} & 0.95 &       \textbf{0.97} &    0.94 & 0.96 &       0.96 &    0.96 & \textbf{0.99} &       \textbf{0.99} &    \textbf{1.00} \\\bottomrule
\end{tabular}\\
\raggedright{\footnotesize \dag\ We excluded results from the SetFit model on the full Enron training set because it did not achieve a meaningful result after 104 hours of training.}\\
\end{table}

We found that large language models outperformed baseline models on the SMS and Enron datasets. However, we also observed that LLMs did not perform better than baseline models on the Ling and SpamAssassin datasets. Among all the datasets, the SMS dataset showed the most significant difference between the best-performing baseline model and the worst-performing large language model in terms of F1 score, with a difference of 0.05 points.

Our experiments show that the Spam-T5 model had the best overall performance, with an average F1 score of 0.9742. The RoBERTa and SetFit models also surpassed the baseline models with the same score of 0.9670. Among the baseline models, the SVM approach performed the best, achieving an average F1 score of 0.9560. Conversely, the XGBoost model was the least effective, with an average F1 score of 0.8842. These outcomes indicate that LLMs are superior to traditional machine learning algorithms in most spam detection scenarios.

\subsection{Few-shot Learning}

In the few-shot learning setting, we evaluated the performance of each model after being trained on $k \in \{4,8,16,32,64,128,256,\text{Full}\}$ samples. The results of our analysis are presented in Table~\ref{tab:results_few}.

\begin{table}[ht]
\centering
\caption{Test F1 score for each model using different numbers of training samples (macro average using the four datasets). The ``Full" column corresponds to the complete training sets (80\% of the original datasets).}\label{tab:results_few}
\begin{tabular}{
p{2.0cm}
>{\centering\arraybackslash}p{1.1cm}
>{\centering\arraybackslash}p{1.1cm}
>{\centering\arraybackslash}p{1.1cm}
>{\centering\arraybackslash}p{1.1cm}
>{\centering\arraybackslash}p{1.1cm}
>{\centering\arraybackslash}p{1.1cm}
>{\centering\arraybackslash}p{1.1cm}
>{\centering\arraybackslash}p{1.1cm}
>{\centering\arraybackslash}p{1.1cm}}
\toprule
\textbf{} & \multicolumn{8}{c}{\textbf{Number of training samples}} \\ \cmidrule(l){2-9} 
\textbf{Model} & \textbf{4} & \textbf{8} & \textbf{16} & \textbf{32} & \textbf{64} & \textbf{128} & \textbf{256} & \textbf{Full} \\ \midrule
NB           & 0.145 & 0.210 & 0.211 & 0.243 & 0.361 & 0.505 & 0.663 & 0.930 \\
LR           & 0.153 & 0.195 & 0.210 & 0.248 & 0.353 & 0.420 & 0.599 & 0.927 \\
KNN          & 0.516 & 0.523 & 0.596 & 0.591 & 0.603 & 0.688 & 0.733 & 0.887 \\
SVM          & 0.155 & 0.267 & 0.288 & 0.334 & 0.531 & 0.732 & 0.858 & 0.952 \\
XGBoost      & 0.000 & 0.079 & 0.351 & 0.431 & 0.600 & 0.666 & 0.767 & 0.877 \\
LightGBM     & 0.000 & 0.000 & 0.000 & 0.000 & 0.455 & 0.608 & 0.703 & 0.948 \\ \midrule
RoBERTa      & 0.241 & 0.174 & 0.575 & 0.738 & 0.459 & 0.915 & 0.929 & 0.970 \\
SetFit       & 0.215 & 0.339 & 0.557 & \textbf{0.855} & \textbf{0.887} & \textbf{0.929} & \textbf{0.941} & 0.967 \\
Spam-T5      & \textbf{0.544} & \textbf{0.534} & \textbf{0.619} & 0.726 & 0.806 & 0.864 & 0.933 & \textbf{0.974} \\ \bottomrule
\end{tabular}\\
\end{table}

The performance profiles of NB, LR, and SVM are similar, with a noticeable improvement on larger datasets. In contrast, KNN achieves relatively higher F1 scores for smaller training sets, with scores exceeding 0.5 for sizes of 4 and 8. However, its performance plateaus as the number of shots increases, with a maximum score of 0.887 on the full datasets. Gradient-boosted tree models, such as XGBoost and LightGBM, exhibit underwhelming results on the smallest dataset sizes (4, 8, 16, and 32). Their performance rapidly improves with an increase in the training set size, culminating in scores of 0.877 and 0.948 on the full datasets, respectively.

RoBERTa's performance is somewhat inconsistent across training set sizes, starting at 0.241 for size 4, dropping to 0.174 for size 8, and then increasing to 0.970 on the full datasets. In contrast, SetFit consistently improves in performance as the number of samples increases, achieving an F1 score of 0.967 on the full datasets. This model performs best on dataset sizes 32, 64, 128, and 256, indicating that SetFit is more suitable for this particular type of ``medium" few-shot learning. Spam-T5, on the other hand, is the best-performing model in very-few-shot scenarios, \ie, when there are only 4--16 samples available for training. Its performance steadily increases with more samples, achieving the highest F1 score on the full datasets.

\begin{figure}[ht]
\begin{minipage}{\textwidth}
\begin{minipage}[c]{0.49\textwidth}
\centering
\includegraphics[width=6cm]{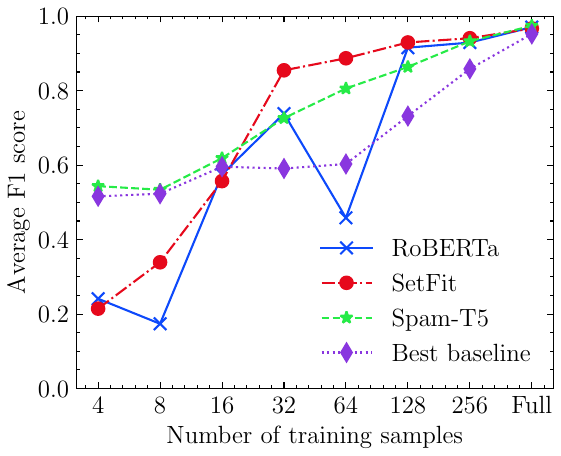}
\captionof{figure}{Comparison of test F1 scores achieved by LLMs vs. the best baseline model for every number of training samples (macro average using four datasets).}\label{fig:errorbars_llm_f1}
\end{minipage}
\hfill
\begin{minipage}[c]{0.49\textwidth}
\centering
\begin{tabular}[b]{@{}l
>{\centering\arraybackslash}p{2.5cm}@{}}
\toprule
\textbf{Model} & \textbf{F1 score} \\ \midrule
NB             & $0.4085_{.2734}$                    \\
LR             & $0.3880_{.2621}$                    \\
KNN            & $0.6421_{.1234}$                    \\
SVM            & $0.5146_{.3005}$                    \\
XGBoost        & $0.4716_{.3159}$                    \\
LightGBM       & $0.3392_{.3871}$                    \\ \midrule
RoBERTa        & $0.6253_{.3139}$                    \\
SetFit         & $0.7112_{.2990}$                    \\
\textbf{Spam-T5}        & \textbf{0.7498$_{.1718}$}           \\ \bottomrule
\end{tabular}
\captionof{table}{Mean test F1 scores and standard deviations across all numbers of training samples (macro average using four datasets).}\label{tab:results_average}
\end{minipage}
\end{minipage}
\end{figure}

Figure~\ref{fig:errorbars_llm_f1} illustrates the consistent superiority of LLMs over the baseline models in terms of F1 score across all training sample sizes. Furthermore, Table~\ref{tab:results_average} presents the mean F1 scores of every model, and shows that Spam-T5 achieves the highest overall performance with an F1 score of 0.7498. These results can be attributed to Spam-T5's high accuracy in the very-few-shot setting and consistent robustness across all sample sizes.

\section{Discussion}

The results of our experiments provide insights into the strengths and limitations of LLMs for spam detection. Specifically, our findings suggest that LLMs, such as RoBERTa and MPNet (model used by SetFit), perform well in general but are outperformed by Spam-T5 in the very-few-shot setting. This difference in performance can be attributed to the number of parameters, with Spam-T5 having 250M parameters compared to RoBERTa's 125M and MPNet's 110M.

\begin{figure}[ht]
\centering
\includegraphics[width=6cm]{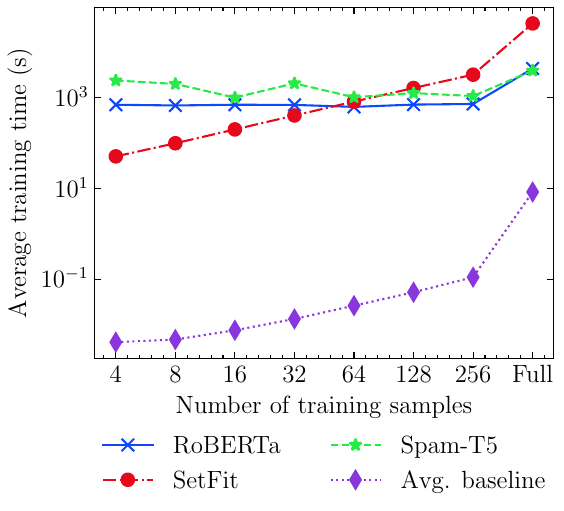}
\hfill
\includegraphics[width=6cm]{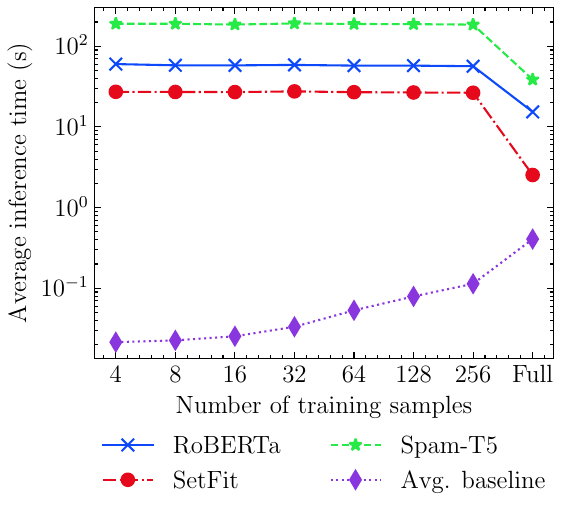}
\caption{Training and inference times for the three LLMs and the average times for the baseline techniques (macro average using the four datasets). Here, ``inference time" corresponds to the execution time to process the entire test set. We exclude training and inference times from the SetFit model on the full Enron dataset because it did not achieve a meaningful score after 104 hours of training.} \label{fig:runtime}
\end{figure}

Moreover, our results indicate a clear trade-off between accuracy and runtime when using LLMs and baseline techniques for spam detection, as illustrated in Figure~\ref{fig:runtime}. While LLMs are more robust and perform better in most cases, they require long training and inference times. In contrast, baseline techniques are faster but do not obtain the same level of accuracy. This suggests that LLMs achieve improved sample efficiency at the expense of computational efficiency. This trade-off highlights the need to consider the specific requirements of the task, such as the available computational resources and the desired level of accuracy.

Figure~\ref{fig:runtime} also shows a surprising increase in inference time for the baseline models as the number of training samples increases (\ie, the number of test samples decreases). This trend is counterintuitive, as we would expect a similar trend to that of the LLMs, where the inference time decreases since there are fewer samples to process. This is due to the sigmoid kernel function used by the SVM model. As the number of training samples increases, the sigmoid kernel requires more computational effort during the inference phase, leading to the observed increase in inference time.

The practical application of LLMs is hindered by their substantial computational requirements for training and deployment, which necessitate the use of specialized hardware, including GPUs and TPUs. Addressing this limitation is an active area of research, with numerous techniques that reduce the memory footprint and computational resources required by LLMs. For instance, some approaches have explored the use of 8-bit floating-point formats to reduce the memory requirements, as demonstrated in \cite{dettmers2022llmint8}. Other methods, such as LoRA \cite{hu2021lora}, aim to reduce the computational resources required to train and deploy LLMs.


\section{Conclusion}

Our study demonstrates the effectiveness of LLMs for email spam detection, particularly in few-shot scenarios. Experiments show that LLMs outperform well-established baseline techniques in terms of F1 score across all training sample sizes. Furthermore, our solution, Spam-T5, achieves the highest overall performance with an average F1 score of 0.7498.

These findings suggest that LLMs, and specifically Spam-T5, could be a valuable tool for addressing the ongoing challenges in spam detection. By incorporating a limited number of fraudulent samples, we can update models and enhance their performance without the need for extensive data labeling efforts. This approach simplifies the task of building robust models that can handle dynamic data distributions, thus offering a practical and effective solution to real-world problems.

In order to deploy LLMs in real-world applications, future work will need to focus on reducing the computational requirements of these models. One approach to achieving this goal involves developing techniques that minimize the memory footprint and computational resources required by LLMs, such as those explored in recent studies.


\section{Ethical Statement}

As spam detection is an essential component of email systems and other communication platforms, using effective language models in this domain can lead to more accurate filtering and improved user experience. However, our research raises ethical concerns about the potential misuse of such language models for censorship. The ability to classify messages as spam or non-spam could be used to suppress or filter out content that does not align with certain political or social agendas. We recognize the importance of safeguarding against such misuse and promoting responsible use of machine learning tools in the public domain.

Furthermore, the development and deployment of large language models have a significant environmental impact, consuming significant amounts of energy and contributing to carbon emissions. We acknowledge the potential ecological consequences of our research and call for greater attention to the environmental sustainability of machine learning models and their applications.


%
%
%
\bibliographystyle{splncs04}
\bibliography{mybibliography}





\end{document}